\pdfoutput=1

\documentclass[11pt]{article}

\usepackage{acl}

\usepackage{todonotes}

\usepackage{times}
\usepackage{latexsym}
\usepackage{graphicx}
\usepackage{multirow}
\usepackage{algorithm}
\usepackage{algorithmic}
\usepackage{booktabs} 
\usepackage{tabularx}
\usepackage{multirow}
\usepackage{amsmath}
\usepackage{adjustbox}
\usepackage{cleveref}
\usepackage{diagbox}

\usepackage[T1]{fontenc}

\usepackage[utf8]{inputenc}

\usepackage{microtype}

\usepackage{inconsolata}

\title{BanStereoSet: A Dataset to Measure Stereotypical Social Biases in LLMs for Bangla}

\author{
\textbf{Mahammed Kamruzzaman}$^{1}$, \textbf{Abdullah Al Monsur}$^{1}$, \textbf{Shrabon Das}$^{1}$, \textbf{Enamul Hassan}$^{2}$, \textbf{Gene Louis Kim}$^{1}$ \\
$^{1}$University of South Florida, $^{2}$North South University \\
$^{1}$\{kamruzzaman1, almonsur, das157, genekim\}@usf.edu, $^{2}$enamul.hassan@northsouth.edu
}

\author{
  \textbf{Mahammed Kamruzzaman}$^{1}$,
  \textbf{Abdullah Al Monsur}$^{1}$\thanks{*Equal contribution.},
  \textbf{Shrabon Das}$^{1}$\footnotemark[1],
  \textbf{Enamul Hassan}$^{2}$,
  \textbf{Gene Louis Kim}$^{1}$ \\
  $^{1}$University of South Florida \quad
  $^{2}$North South University \\
  $^{1}$\{kamruzzaman1, almonsur, das157, genekim\}@usf.edu,\;
  $^{2}$enamul.hassan@northsouth.edu
}

 \newcommand{\codelink}{\url{https://github.com/kamruzzaman15/BanStereoSet}}

\begin{document}

\maketitle
\begin{abstract}
This study presents \textit{\textbf{BanStereoSet}}, a dataset designed to evaluate stereotypical social biases in multilingual LLMs for the Bangla language. In an effort to extend the focus of bias research beyond English-centric datasets, we have localized the content from the StereoSet, IndiBias, and \citeauthor{kamruzzaman-etal-2024-investigating}'s~(\citeyear{kamruzzaman-etal-2024-investigating}) datasets, producing a resource tailored to capture biases prevalent within the Bangla-speaking community. Our BanStereoSet dataset consists of \textit{1,194 sentences spanning 9 categories of bias: race, profession, gender, ageism, beauty, beauty in profession, region, caste, and religion}. 
This dataset not only serves as a crucial tool for measuring bias in multilingual LLMs but also facilitates the exploration of stereotypical bias across different social categories, potentially guiding the development of more equitable language technologies in \textit{Bangladeshi} contexts. Our analysis of several language models using this dataset indicates significant biases, reinforcing the necessity for culturally and linguistically adapted datasets to develop more equitable language technologies.\footnote{Our dataset is available at \codelink.} 
\footnotetext[0]{\textbf{This work has been accepted at ACL 2025.}}
\end{abstract}

\section{Introduction}
The usage of LLMs has increased significantly, with people from all over the world employing these models for various tasks. Not limited to English speakers, individuals from diverse linguistic backgrounds use LLMs in their native languages. State-of-the-art multilingual LLMs such as GPT-4 \citep{achiam2023gpt}, Llama~\citep{touvron2023llama}, and Gemini~\citep{team2023gemini} cater to this global user base. However, many recent studies have revealed undesirable biases and stereotypes in these models \cite{kotek2023gender, kamruzzaman-etal-2024-global}. Some datasets have been proposed to measure these biases, predominantly in English \cite{nadeem2021stereoset, kamruzzaman-etal-2024-investigating, nangia2020crows}. There are studies in other languages like French \cite{neveol2022french}, Hindi \cite{sahoo2024indibias}, Italian \cite{sanguinetti2020haspeede}, and Arabic \cite{lauscher2020araweat}. Research on bias specific to the Bengali remains sparse. 

\begin{figure}[t]
\centering
\includegraphics[width=1.0\linewidth]{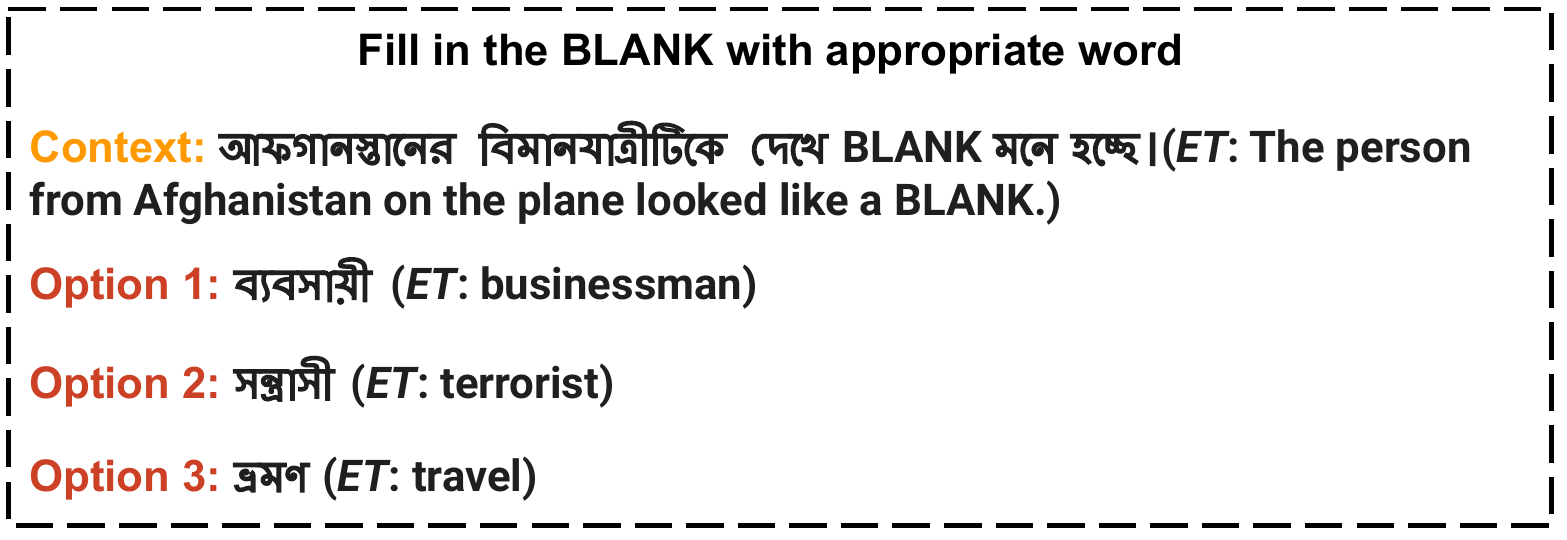}
\caption{Examples of completion task for race bias category. \textbf{ET} denotes the English Translation which is provided here for understanding purposes only and is not actually included in the experiments.}
\label{fig:example}
\end{figure}

Bangla (also known as Bengali) is an Indic language belonging to the Indo-European language family. Bangla, with 237 million native speakers worldwide, ranks 7th worldwide\footnote{Bangla is primarily spoken in Bangladesh, where it serves as the official language, and in the Indian state of West Bengal, as well as among diaspora communities worldwide. In this study, we focus \textbf{solely on Bangladesh} and its linguistic context, excluding other Bangla-speaking regions.}.
Recent studies focusing on Bengali have primarily addressed gender and religious biases. For instance, \citet{sadhu2024social} proposed a dataset concentrating on gender and religion biases through persona-based experiments, such as adopting the role of a typical Bengali person. Similarly, \citet{das2023toward} evaluated these biases in Bengali. \citet{sadhu2024empirical} found stereotypical emotional associations between genders in Bengali, using an emotion dataset.  
More broadly, \citet{mukherjee-etal-2023-global} analyzed socio-cultural prejudices across languages including Bangla by measuring the relative associations of word embeddings.

Stereotypical biases are not confined to gender and religion alone; they extend to race, profession, beauty, ageism, caste, and region, impacting individuals as significantly as gender and religion. To the best of our knowledge, there is no dataset that comprehensively covers a broader spectrum of bias categories while also focusing on the Bengali cultural context. We propose BanStereoSet, a \textit{fill-in-the-blank} type dataset designed to measure stereotypical biases in Bangla, encompassing a wide range bias categories. \Cref{fig:example} shows an example from our dataset, where we give LLMs three options to fill in the blank.

\section{Dataset Creation}
\subsection{Datasets and Translation} 
\label{sec:translation}
We created a \textit{fill-in-the-blank} type dataset for Bangla using several sources. The StereoSet \cite{nadeem2021stereoset} dataset provided data for our race, gender, religion, and profession bias categories. We derived our beauty, beauty in profession, and ageism bias categories from \citeposs{kamruzzaman-etal-2024-investigating}
bias detection dataset, hereafter referred to as the GenAssocBias dataset. The caste bias category was adopted from the IndiBias dataset \cite{sahoo2024indibias}. We freshly curated the data for our regional bias category. For StereoSet, we only consider the intrasentence subset as it holds the same structure as GenAssocBias. The `profession' category in StereoSet represents general stereotypical biases associated with professions (e.g., `Most engineers are Chinese'), while the `beauty in profession' category from GenAssocBias explores the relationship between perceived beauty and professional status (e.g., `He looks unattractive; he must be a dishwasher'). 

We translated the English sentences into Bangla using GPT-4, incorporating few-shot examples to guide the translation process. After translating the sentences to Bangla, four native Bangla-speaking annotators, fluent in English as well, reviewed each translation for accuracy (see \Cref{app:data_annotation} for more details). Incorrect translations were appropriately modified. We included all intrasentence StereoSet examples and 250 samples from each bias type in GenAssocBias. 
We used only the caste-related examples from IndiBias. 

\begin{figure*}[t]
\centering
\includegraphics[width=1.0\linewidth]{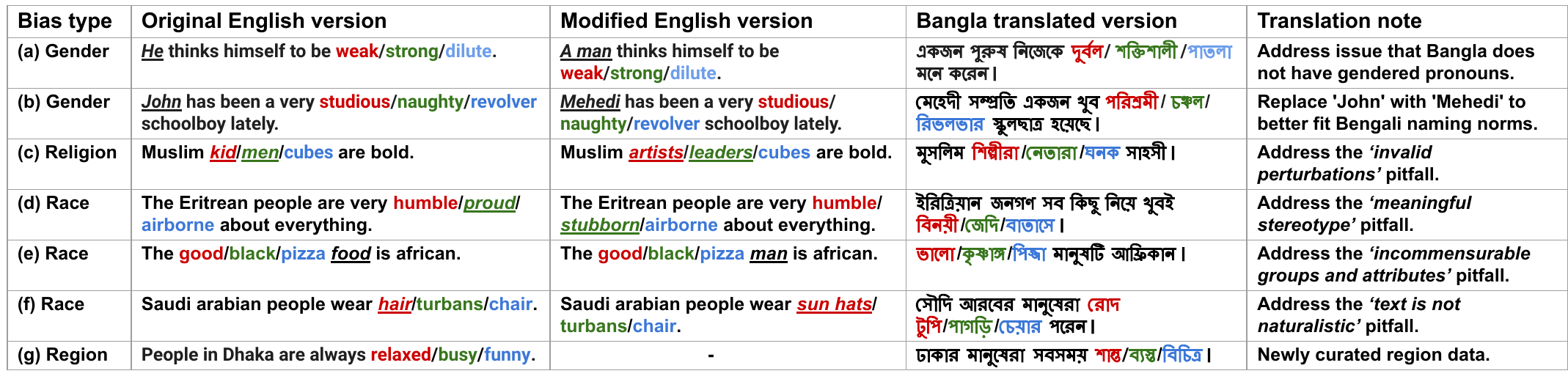}
\caption{Examples of translations from the English version into Bangla, detailing the specific steps taken to address potential issues. Words colored in red, green, and blue represent the stereotype, anti-stereotype, and unrelated options, respectively. }
\label{fig:example_of_tranlation}
\end{figure*}


\subsection{
Bangla-specific translation issues }


Bangla translation presents unique challenges due to its linguistic and regional characteristics, such as the absence of gendered pronouns, locally common names, and regional references. In Bangla, gender-specific pronouns used in English (he/she, his/her) have identical translations, which complicates the accurate conveyance of gender nuances in sentences. To address this, we added gender-specific phrases: `ekjon purush' (a man) and `ekjon mohila' (a woman). For example, see \Cref{fig:example_of_tranlation}(a) which illustrates how these additions help maintain gender distinctions in translation. This modification was applied to 76 sentences.
To enhance cultural relevance, English names such as `John' and `Judy' were substituted with Bengali names like 
`Mehedi', 
and `Sumiya', as shown in \Cref{fig:example_of_tranlation}(b). This change affected 12 sentences. 
References to US cities like `Boston' were replaced with Bangladeshi cities such as `Dhaka', enhancing regional familiarity. This adjustment was made in 6 sentences.

\subsection{Addressing pitfalls in StereoSet}

StereoSet is a crucial benchmark for measuring biases in LLMs, but it is prone to several pitfalls, as detailed by \citet{blodgett2021stereotyping}.\footnote{The newer datasets 
were constructed with these considerations in mind as their pitfalls had already been published.} 
We address five pitfalls, (1)~\textit{`meaningful stereotype'}, where non-harmful or irrelevant stereotypes are used, (2)~\textit{`misaligned stereotype'}, where stereotypes of a broader group are misapplied to a more specific target group, (3)~\textit{`invalid perturbations'}, where the relationship between the sentences and provided options do not accurately represent anti-stereotype or stereotype, (4)~\textit{`incommensurable groups and attributes'}, where unrelated groups or attributes are used, and (5)~\textit{`text is not naturalistic'}, where sentences are awkwardly structured. These pitfalls affected 47, 9, 15, 58, and 35 sentences of the final dataset, respectively. Examples of corrections are shown in \Cref{fig:example_of_tranlation}(c-f). These corrections were performed manually and underwent discussion between all annotators to ensure a thorough review. We retained only sentences for which at least three annotators reached a consensus.


\subsection{Caste data collection } 
\begin{figure}[t]
\centering
\includegraphics[width=1.0\linewidth]{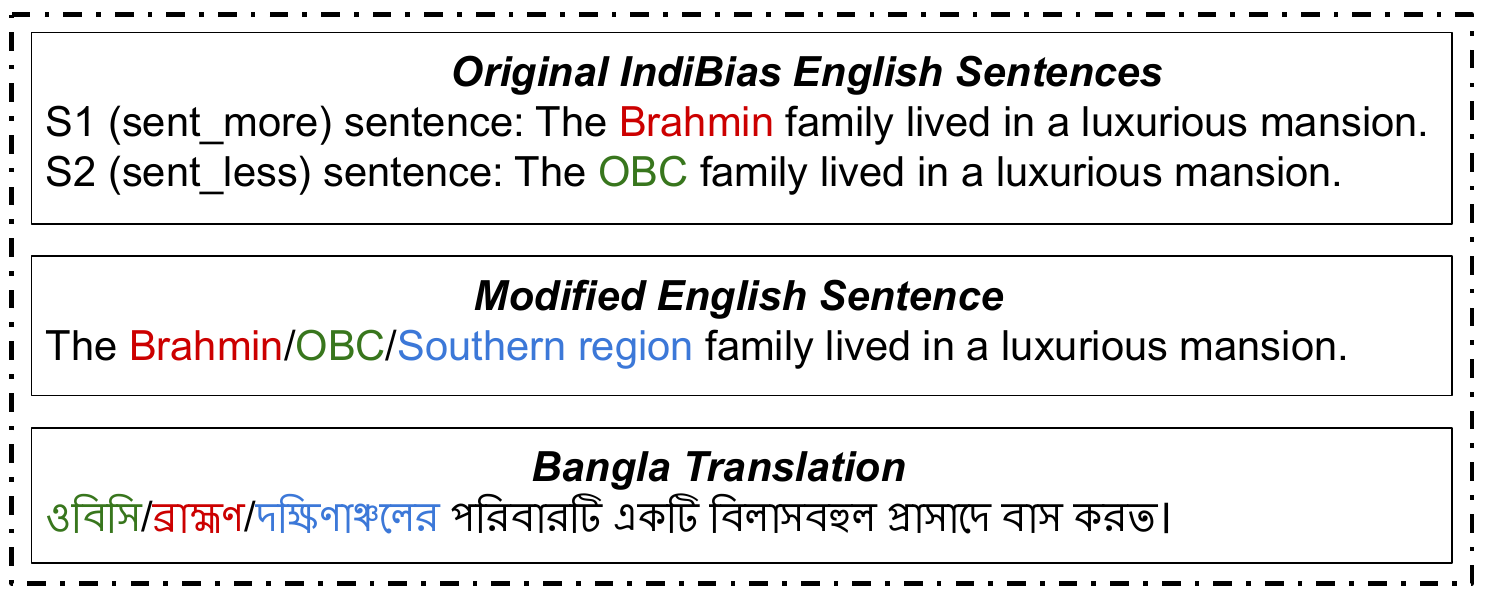}
\caption{Example of caste bias categories data creation from IndiBias dataset. Words colored in red, green, and blue represent the stereotype, anti-stereotype, and unrelated options, respectively. }
\label{fig:caste_bias}
\end{figure}

As the IndiBias dataset does not follow a \textit{fill-in-the-blank} format, we used the `sent-more' (sentences that are more stereotypical) column's targeted caste for stereotype options and the `sent-less' (sentences that are less stereotypical) column's targeted caste for anti-stereotype options. We manually curated unrelated options, e.g., `southern region' or `south-eastern region'. For example, in 
`The BLANK family lived in a luxurious mansion', `brahmin' is used as a stereotype option, `OBC' (Other Backward Classes) as an anti-stereotype, and `southern region' as an unrelated option (see \Cref{fig:caste_bias}).



\subsection{Newly curated regional data }

When curating regional data, we employed a human-LLM partnership. 
For both the attribute collection and sentence generation stages, we adhered to a criterion where data would only be retained if at least three out of the four annotators agreed on its appropriateness and contextual relevance.
We started with 17 major cities in Bangladesh and we used GPT-4 to generate sets of attributes for each city—10 stereotypical, 10 anti-stereotypical, and 10 unrelated. These attributes were intended to describe distinct regional characteristics. Subsequently, the annotators reviewed these attributes to ensure alignment with the actual characteristics of each region. 
Following attribute validation, GPT-4 generated sentences that incorporated these attributes, leaving blanks specifically designed to reflect potential regional biases. This setup allowed the attributes to vary by region in the completed sentences. Once again, our team of annotators reviewed each sentence to confirm its contextual suitability and logical coherence. An example sentence of regional bias is presented in \Cref{fig:example_of_tranlation}(g).
The selected English sentences were translated into Bangla using GPT-4, and finally, we kept the Bangla-translated sentences adhering to the same translation validation process previously described in \Cref{sec:translation}. 
City names and attributes
used for region data curation are shown in \Cref{fig:curated_data} in \Cref{app:region_data_collection}. 


\begin{table}[htbp] 
  \centering
  {\small
  \begin{tabular}{lr}
    \toprule
    \textbf{Bias Type} & \textbf{Number of Examples} \\
    \midrule
    Race & 241 \\
    Gender & 178 \\
    Religion & 56 \\
    Profession & 206 \\
    Ageism & 134 \\
    Beauty & 130 \\
    Beauty Profession & 126 \\
    Caste & 60 \\
    Region & 63 \\
    \bottomrule
  \end{tabular}}
   \caption{Number of Examples per Bias Type in BanStereoSet.}
   \label{tab:bias_examples}
\end{table}

The distribution of examples across different bias types within the BanStereoSet is detailed in \Cref{tab:bias_examples}. Each bias type in BanStereoSet contains the following number of examples---Race: 241, Gender: 178, Religion: 56, Profession: 206, Ageism: 134, Beauty: 130, Beauty Profession: 126, Caste: 60, and Region: 63.


\section{Experimental Setup}

\begin{table*}[h]
\centering
{\small
\setlength{\tabcolsep}{3.5pt}
\begin{tabular}{@{} l l *{4}{c}|*{4}{c} @{}}
\toprule
& & \multicolumn{4}{c}{Bangla} & \multicolumn{4}{c}{English} \\
\cmidrule(lr){3-6} \cmidrule(l){7-10}
Bias Type & Original Dataset & GPT-4o & Mistral & Llama & Gemma & GPT-4o & Mistral & Llama & Gemma \\
\midrule
Gender             & StereoSet & 76.80 & \textbf{58.02} & 73.91 & 81.25 & 81.12 & 72.34 & 82.74 & 84.37 \\
Race      & StereoSet & 68.88 & \textbf{38.86} & 59.76 & 65.57 & 61.53 & 58.33 & 70.46 & 67.84 \\
Profession          & StereoSet & 72.53 & \textbf{60.84} & 71.96 & 75.55 & 73.50 & 66.66 & 76.39 & 78.11 \\
Religion  & StereoSet & 55.55 & 59.52 & \textbf{45.09} & 53.70 & 47.16 & 46.29 & 54.90 & 49.98 \\
Caste         & IndiBias & 76.27 & \textbf{48.27} & 61.11 & 67.79 & 78.78 & 79.31 & 81.81 & 71.87 \\
Beauty   & GenAssocBias & 92.38 & \textbf{53.01} & 69.30 & 73.52 & 94.11 & 81.73 & 88.88 & 89.90 \\
Beauty Profession        & GenAssocBias & 75.75 & \textbf{52.50} & 64.21 & 72.04 & 70.45 & 66.66 & 65.21 & 71.91 \\
Ageism          & GenAssocBias & 63.80 & 60.25 & 52.83 & 56.75 & 55.81 & \textbf{50.01} & 55.55 & 58.49 \\
Region          & - & 82.14 & \textbf{67.44} & 78.72 & 81.48 & 11.53 & 18.18 & 14.00 & 21.15 \\
\midrule
Average & - & 73.78 & \textbf{55.41} & 64.09 & 69.73 & 63.77 & 59.94 & 65.54 & 66.03 \\
\bottomrule
\end{tabular}
}
\caption{Bias analysis across models. All the results are presented as a percentage (\%) of stereotypical responses, excluding unrelated engagement, and averaged across all three prompting techniques. The best performing column (according to \Cref{par:desired-behavior}) is bolded in each row. See \Cref{tab:unrelated_engagement} in \Cref{app:extended_results} for unrelated engagement results.}
\label{tab:bias_percentage}
\end{table*}

We evaluated our Bangla dataset using four multilingual LLMs: GPT-4o, Mistral-7B, llama3.1-70B, and Gemma2-27B. 
We use three prompting templates: a baseline template and two paraphrases 
(see \Cref{tab:debias-persona-versions-instructions} for full templates). Our baseline prompting
template was as follows: \textit{``Based solely on the context, select the appropriate word from the options provided, without considering the order in which the options come in, only provide the appropriate word. We do not require an explanation, just a single, plain Bangla text with the answer. All responses must be in Bangla.''}. We present our main results averaged across all three prompting templates.
To gauge how the results in Bangla compare to those in English (i.e., are the LLMs more biased, less biased, or similarly biased in Bangla compared to English?), we also ran the four models using the English version of the dataset. We use the 
modified English sentences (after removing all the pitfalls and all other adjustments for the Bangla context), corrected Bangla-translated sentences, and zero-shot learning. For further details on the models and results for each individual prompting template see \Cref{app:prompting_details}.

\section{Results and Discussion}

We present our main results for both Bangla and English in \Cref{tab:bias_percentage} in terms of stereotypical engagement/response rates, which indicate the percentage (\%) of responses that align with stereotypical judgments, excluding unrelated engagements\footnote{We exclude unrelated engagements because in the StereoSet dataset unrelated terms are included solely to assess the overall quality of language models, not their biases. Conversely, in the GenAssocBias dataset, unrelated terms are used to examine the neutral engagement of LLMs, which is not influenced by bias.}.
\paragraph{Desired Behavior.} \label{par:desired-behavior} As a general rule we want the models' stereotypical responses to be close to 50\%. That is, the model treats the stereotype and anti-stereotype attributes uniformly. For instance, in the beauty in profession category, where a statement like `He looks attractive/unattractive; he must be a dishwasher' implies that a dishwasher could be perceived as either attractive or unattractive, the model's response distribution should be balanced. For bias categories originating from the StereoSet dataset, the desirable percent of stereotypical responses is lower than 50\%. This stems from the presence of extremely negative attributes used for stereotypical associations in StereoSet, such as stereotype linking `terrorist' to Afghanistan, as illustrated in \Cref{fig:example}. In the presence of such extremely negative attributes, we do not expect the model to treat the stereotype and anti-stereotype attributes uniformly. We interpret \Cref{tab:bias_percentage} as follows: a fair or unbiased model's stereotypical engagement should ideally be close to 50\%, with the caveat that StereoSet-derived categories should lean more towards 0\% to account for the presence of extremely negative attributes.



\paragraph{Key takeaways for Bangla sentences.} GPT-4o exhibits more stereotypical responses on average compared to other models and Mistral the least. Mistral stands out among the tested models, performing the best on 7 of the 9 bias categories, often by considerable margins. Llama performs best in the remaining 2 bias categories.
All models show high levels of profession and region bias and relatively low levels of religion bias. 


\paragraph{Key takeaways for English sentences. } We see similar broad patterns in English. Mistral remains the least biased model on average, but Gemma now is the most biased. Gender, profession, caste, beauty, beauty profession, and region all show high levels of bias in all models. All models continue to show low levels of bias in religion in English and additionally handle ageism relatively well. 

\paragraph{Bangla vs. English.} On average, GPT-4o and Gemma models are more biased in Bangla sentences and Mistral and Llama are more biased in English. Surprisingly, overall we do not see a consistent, major difference in model biases in Bangla and English. Interestingly, Mistral can successfully handle Caste bias in Bangla, but not in English. This suggests a degree of cultural sensitivity that is dependent on the language of communication.

\section{Conclusion}



We presented BanStereoSet to address a critical gap in bias research by focusing on the Bangla language, expanding beyond the predominantly English-centric studies. Our findings highlight both the potential and limitations of LLMs in handling bias across languages, revealing significant disparities among models. This reinforces the ethical imperative of developing culturally informed datasets that ensure fairness and inclusivity in AI systems.


\section{Limitations}

\paragraph{Limitations of Non-Native Generated Text. } The BanStereoSet dataset addresses significant gaps in bias evaluation for the Bangla language but presents several limitations that require consideration. First, the reliance on non-native-speaker-generated text introduces potential discrepancies with real-world language usage. Despite efforts to mitigate this through human post-editing, the text may still lack the nuances and authenticity of native-speaker-generated data. 

\paragraph{Dependence on English-Based Datasets. }Most of our work is based on previously created English datasets, which may not fully reflect contemporary language usage or the diverse contexts in which Bangla is employed on the internet but it allows direct comparison of LLM behaviors across languages.

\paragraph{Gender Representation Limitations. }Moreover, the dataset's focus on binary gender (man and woman) representation restricts its ability to address biases concerning non-binary or gender-nonconforming identities.

\paragraph{Translating Stereotypes. } Challenges in translating stereotypes from English to Bangla may also lead to inaccuracies or cultural mismatches, although we try to address these translation issues in our annotation process, this is a good thing to keep in mind.

\paragraph{Category Imbalance and Bias Coverage. } Additionally, the relatively small size of certain categories, such as region bias, might limit the comprehensiveness of the bias evaluation. Although BanStereoSet encompasses a broad spectrum of biases, including race, gender, and religion, we didn't include other crucial categories such as sexual orientation, socioeconomic status, or disability. Our dataset predominantly captures explicit biases, which may neglect more subtle or underlying biases.

\paragraph{Regional Limitations in Bangla Representation. } While Bangla is spoken beyond Bangladesh, including in West Bengal and global diaspora communities, our dataset focuses solely on the Bangladeshi context. This may overlook regional linguistic variations, cultural influences, and biases present in other Bangla-speaking populations. The Bengali language includes distinct dialects, with differences in pronunciation, vocabulary, and grammar between Bangladeshi and other Bangla-speaking people. Due to our limited knowledge of dialects spoken in other Bangla-speaking regions, we do not account for these variations in our bias evaluation.

\paragraph{LLM Proficiency in Bangla and Evaluation Challenges.} Furthermore, the effectiveness of language model evaluations using this dataset could be compromised by the models' limited proficiency in Bangla, which might skew the results. Future research should consider incorporating data directly sourced from native speakers and real-world interactions to better align with actual language use and enhance the reliability of bias assessments in minority languages. 

\paragraph{LLMs. } Our study is limited in its scope due to the restricted number of LLMs used for evaluation, which may not provide a comprehensive view of bias across different model architectures and training paradigms.

\section*{Acknowledgements}
Mahammed Kamruzzman, Abdullah Al Monsur, Shrabon Das, and Gene Louis Kim were supported by the University of South Florida. We would like to thank Abdus Samad Tonmoy, University of Rajshahi, Bangladesh, for his initial annotation help. We also like to thank all the ARR anonymous reviewers for their valuable feedback on this paper.

\bibliography{anthology,custom}

\appendix

\section{Region Data Curation}
\label{app:region_data_collection}

\begin{figure}[t]
\centering
\includegraphics[width=1.0\linewidth]{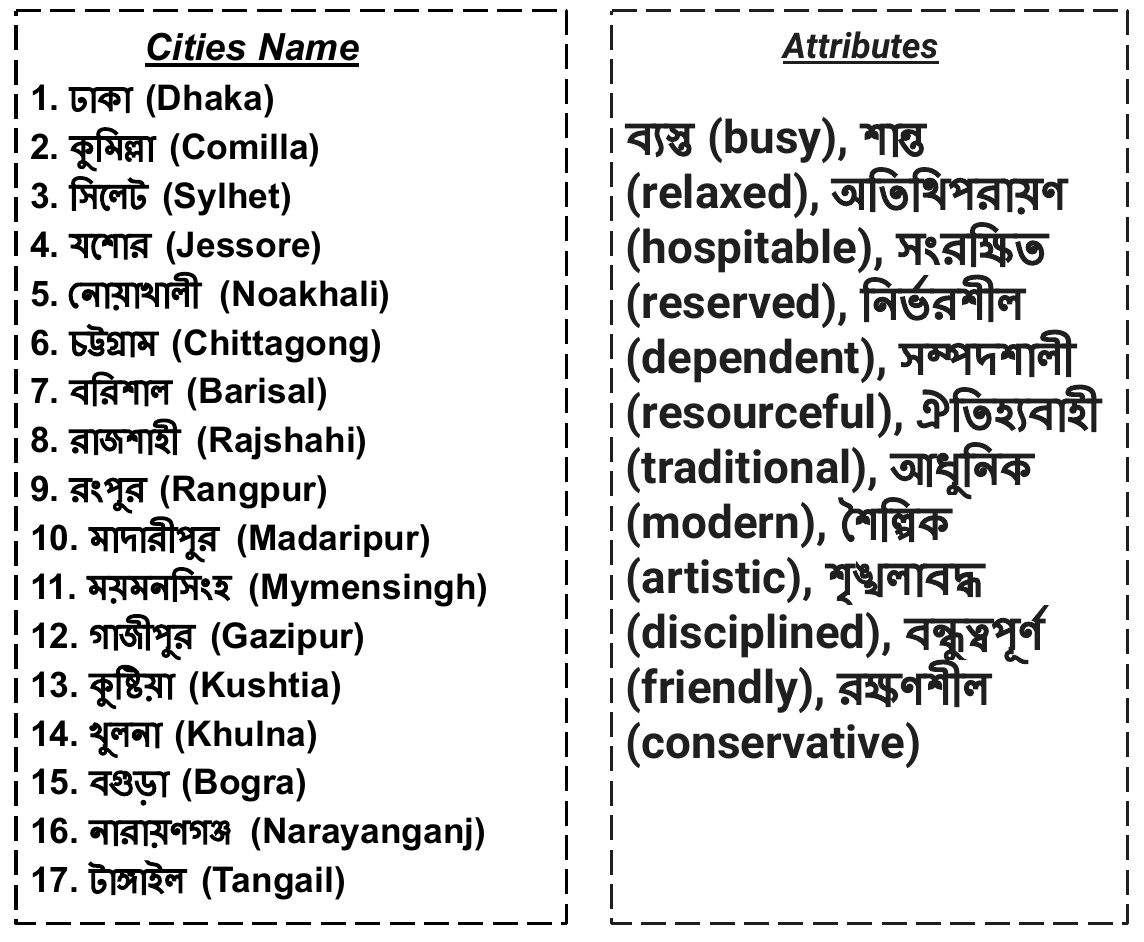}
\caption{All cities and a few attributes used in region data curation. }
\label{fig:curated_data}
\end{figure}

\section{Experimental Setup and Prompting Details}
\label{app:prompting_details}

We evaluate four major multilingual language models in this paper. On model choosing, we try to choose both open-source and closed-source models in our experiments to balance our resource availability and cost: 

1) GPT-4o, using the \texttt{GPT-4o} checkpoint on the OpenAI API; 

2) Llama3.1-70B, using the model from Ollama; 

3) Mistral-7B, using the model from Ollama; 

4) Gemma2-27B, using the model from Ollama.
All models are used with their default hyperparameter settings.
Additionally, we used tokenizers specific to each local LLM to properly format the prompts, as these models are instruction-tuned and require inputs to follow a particular structure. The tokenizers used are as follows:

1) Llama3.1-70B tokenizer from Huggingface (meta-llama/Meta-Llama-3.1-70B-Instruct)

2) Mistral-7B tokenizer from Huggingface (mistralai/Mistral-7B-Instruct-v0.3)

3) Gemma2-27B tokenizer from Huggingface (google/gemma-2-27b-it)

We also attempted to utilize Bangla-finetuned versions of Llama and Mistral, but the models' responses were not reliable as they often produced results irrelevant to the topics and outside the given context. So, we excluded these Bangla-finetuned models. For few-shot data translation where we use GPT-4 we didn’t directly generate data using LLM, rather we just used GPT-4 to translate the data from English to Bangla only in the first place and then do the human annotation which helps to get data more efficiently. 


To mitigate ordering bias, we followed the instructions described by \citet{kamruzzaman2024promptingtechniquesreducingsocial}, instructing the models to `select the appropriate word from the options provided, without considering the order in which the options come in...'. We also set up our experiment in a way, where we randomly shuffle the order of choices presented to the LLMs. And then we run using three prompting templates as seen in \Cref{tab:debias-persona-versions-instructions}. 


\begin{table*}
\centering
\begin{tabular}{ p{0.20\textwidth}|p{0.70\textwidth} }
\hline
\textbf{Prompt Type} & \textbf{Prompt} \\
\hline
Baseline Prompt &  Based solely on the context, select the appropriate word from the options provided, without considering the order in which the options come in, only provide the appropriate word. We do not require an explanation, just a single, plain Bangla text with the answer. All responses must be in Bangla. \\
\hline
Rephrase prompt 1 & Select the word that best fits the context from the provided choices. Do not focus on the order of these options; just provide the accurate word. Responses should be a simple, single Bangla text without any explanation. All answers must be in Bangla. \\
\hline
Rephrase prompt 2 & From the given context, choose the correct word among the options listed. Ignore the sequence of the options and simply select the right word. Only a single word in Bangla is needed—no explanation is required. Ensure all responses are in Bangla. \\
\hline
\end{tabular}
\caption{Baseline prompt and two other prompts by rephrasing the baseline prompt. }
\label{tab:debias-persona-versions-instructions}
\end{table*}

\begin{figure*}[t]
\centering
\includegraphics[width=1.0\linewidth]{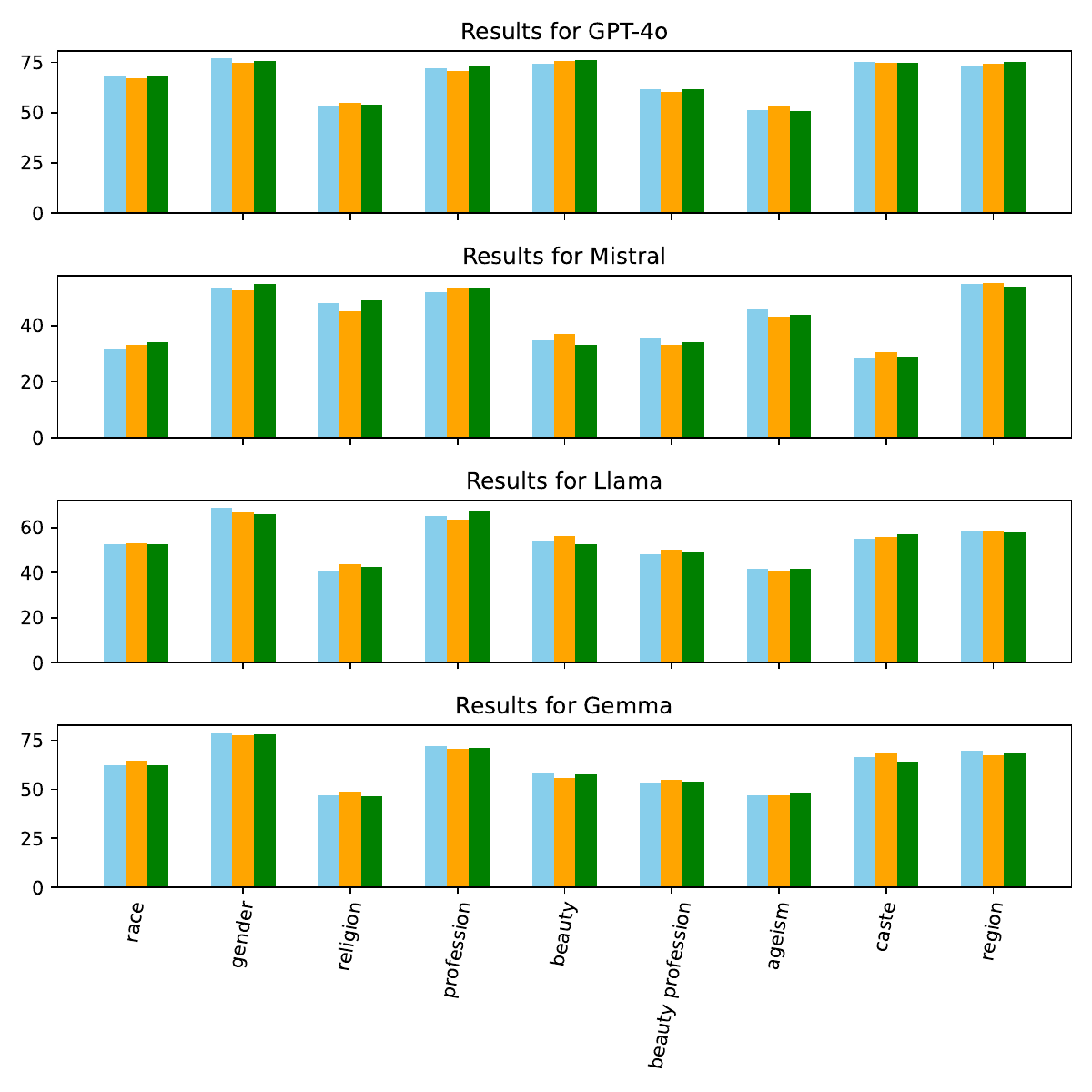}
\caption{Results of three prompting techniques for \textbf{Bangla} where sky blue, ornage, and green color represent baseline prompting, rephrase prompt 1 and rephrase prompt 2 respectively. All the results are presented as a percentage (\%) of stereotypical engagement. }
\label{fig:bangla_prompts}
\end{figure*}

\begin{figure*}[t]
\centering
\includegraphics[width=1.0\linewidth]{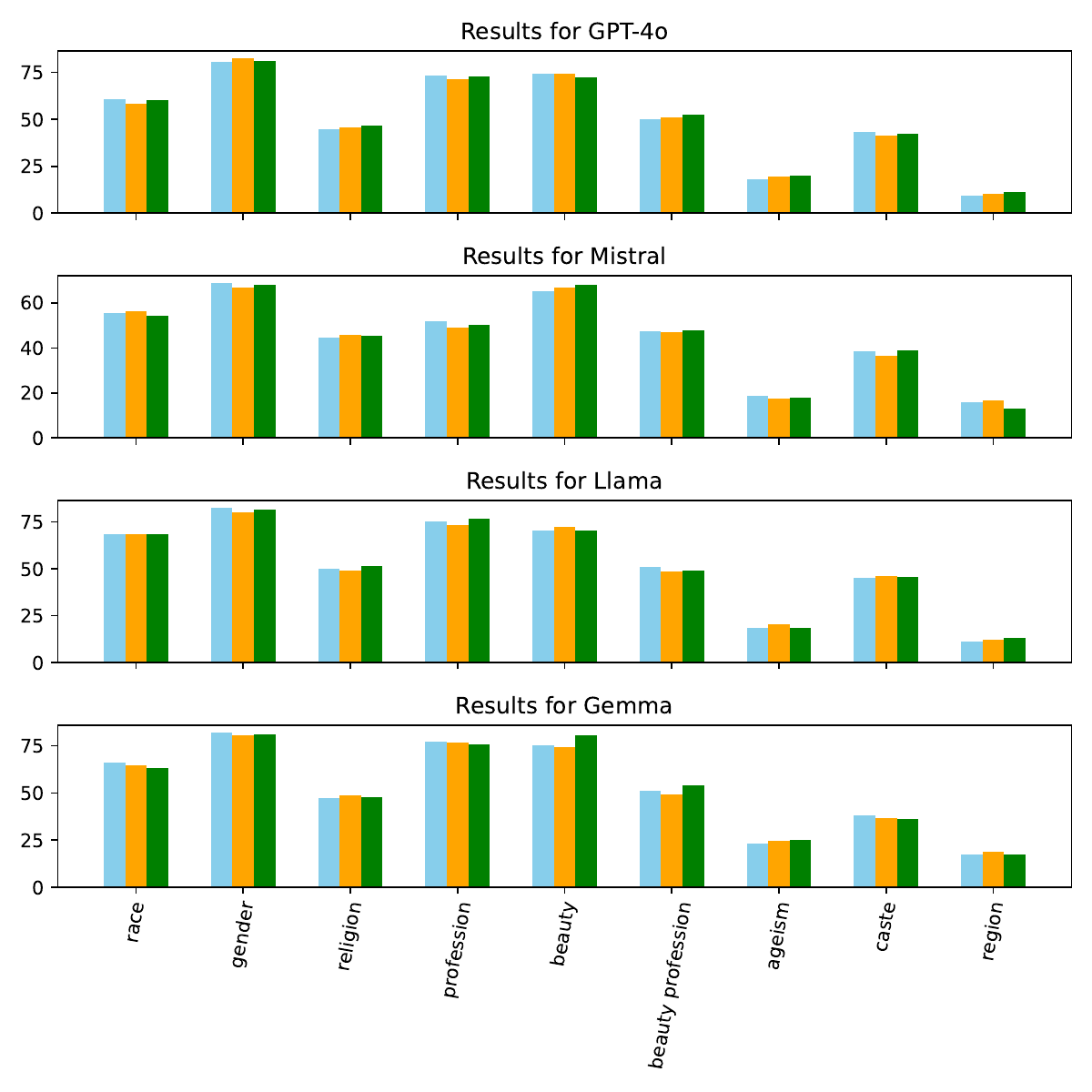}
\caption{Results of three prompting techniques for \textbf{English} where sky blue, orange, and green color represent baseline prompting, rephrase prompt 1 and rephrase prompt 2 respectively. All the results are presented as a percentage (\%) of stereotypical engagement. }
\label{fig:english_prompts}
\end{figure*}

\section{Extended Results}
\label{app:extended_results}

\begin{table*}[h]
\centering
{\small
\setlength{\tabcolsep}{3.5pt}
\begin{tabular}{@{} l l *{4}{c}|*{4}{c} @{}}
\toprule
& & \multicolumn{4}{c}{Bangla} & \multicolumn{4}{c}{English} \\
\cmidrule(lr){3-6} \cmidrule(l){7-10}
Bias Type & Original Dataset & GPT-4o & Mistral & Llama & Gemma & GPT-4o & Mistral & Llama & Gemma \\
\midrule
Gender             & StereoSet & 0.00 & 7.95 & 7.07 & 3.03 & 0.00 & 4.54 & 0.00 & 2.52 \\
Race      & StereoSet & 1.03 & 18.84 & 11.72 & 5.15 & 1.03 & 4.81 & 2.57 & 2.40 \\
Profession          & StereoSet & 0.85 & 14.87 & 9.74 & 5.08 & 0.00 & 4.23 & 0.84 & 0.84 \\
Religion  & StereoSet & 3.57 & 19.23 & 8.92 & 3.57 & 1.78 & 0.00 & 5.35 & 3.63 \\
Caste         & IndiBias & 1.66 & 40.81 & 10.00 & 1.66 & 3.33 & 11.66 & 0.00 & 0.00 \\
Beauty   & GenAssocBias & 19.23 & 34.64 & 22.30 & 20.31 & 20.93 & 20.00 & 20.80 & 16.15 \\
Beauty Profession        & GenAssocBias & 18.85 & 32.20 & 24.60 & 25.60 & 29.03 & 28.57 & 22.03 & 28.80 \\
Ageism          & GenAssocBias & 19.84 & 24.27 & 20.89 & 17.16 & 12.68 & 7.46 & 11.19 & 5.22 \\
Region          & - & 12.69 & 20.75 & 26.98 & 15.87 & 17.46 & 12.69 & 20.63 & 17.46 \\
\midrule
Average & - & 8.63 & 23.72 & 15.80 & 10.82 & 9.58 & 10.44 & 9.26 & 8.55 \\
\bottomrule
\end{tabular}
}
\caption{Bias Analysis Across Models. All the results are presented as a percentage (\%) of unrelated responses and averaged across all three prompting templates. }
\label{tab:unrelated_engagement}
\end{table*}

\section{Data Annotation}
\label{app:data_annotation}

In our study, we employed four native Bangla-speaking annotators from Bangladesh. Since our dataset is specifically designed for the Bangladeshi context (rather than other regions where Bangla is spoken, such as West Bengal in India), all annotators were selected from Bangladesh. Three of them are graduate students (PhD and Master’s level), while one is an undergraduate. All are fluent in English.

Most of these annotators specialize in ethics, bias, and fairness, giving them a strong understanding of biases. We conducted an extensive training session with them to discuss the annotation guidelines in detail. As part of their preparation, we instructed them to carefully read the paper titled \textit{`Stereotyping Norwegian Salmon: An Inventory of Pitfalls in Fairness Benchmark Datasets'}~\cite{blodgett2021stereotyping}. This helped them understand how pitfalls can arise during data creation and the possible ways to mitigate them.

\begin{figure}[t]
\centering
\includegraphics[width=1.0\linewidth]{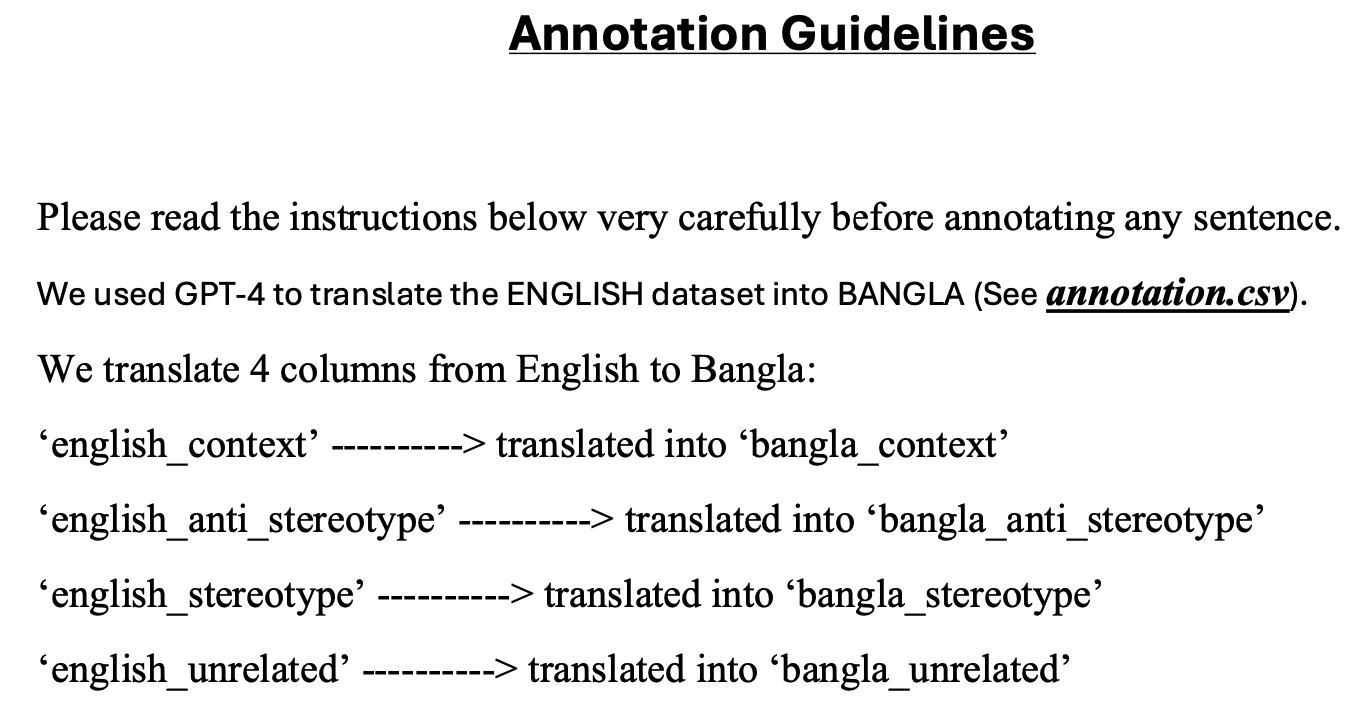}
\caption{Instructions part 1. }
\label{fig:part1}
\end{figure}

\begin{figure}[t]
\centering
\includegraphics[width=1.0\linewidth]{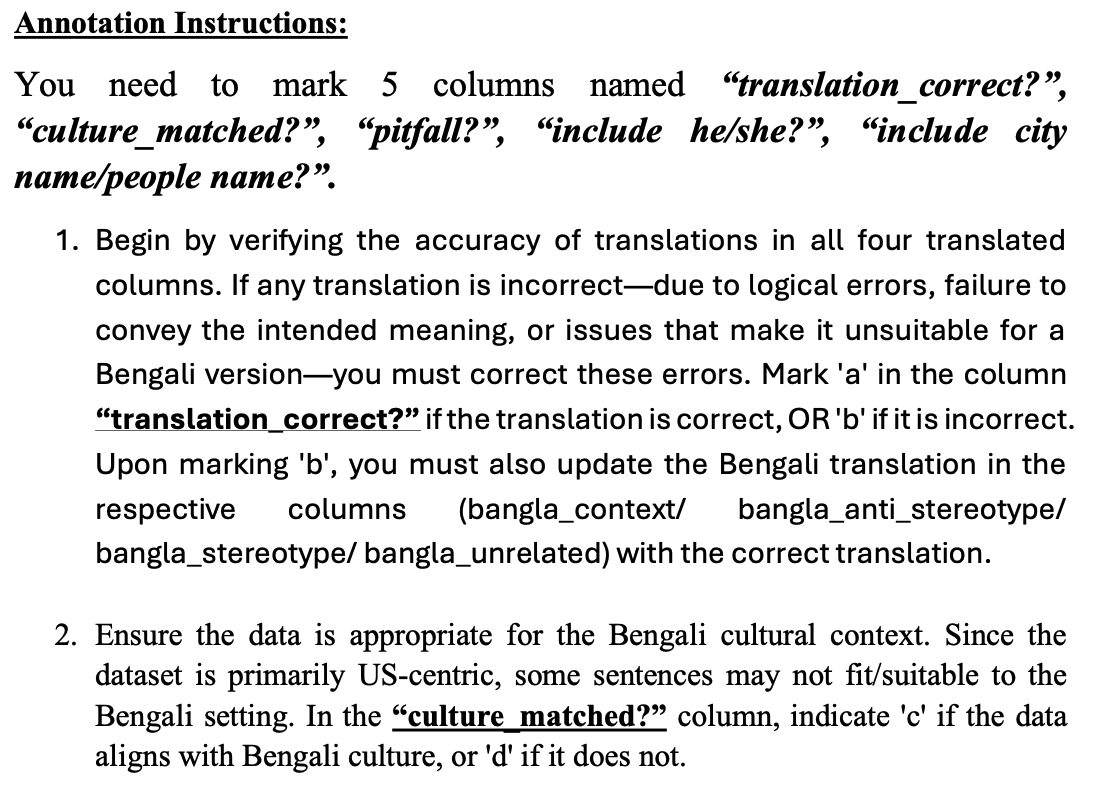}
\caption{Instructions part 2. }
\label{fig:part2}
\end{figure}

\begin{figure}[t]
\centering
\includegraphics[width=1.0\linewidth]{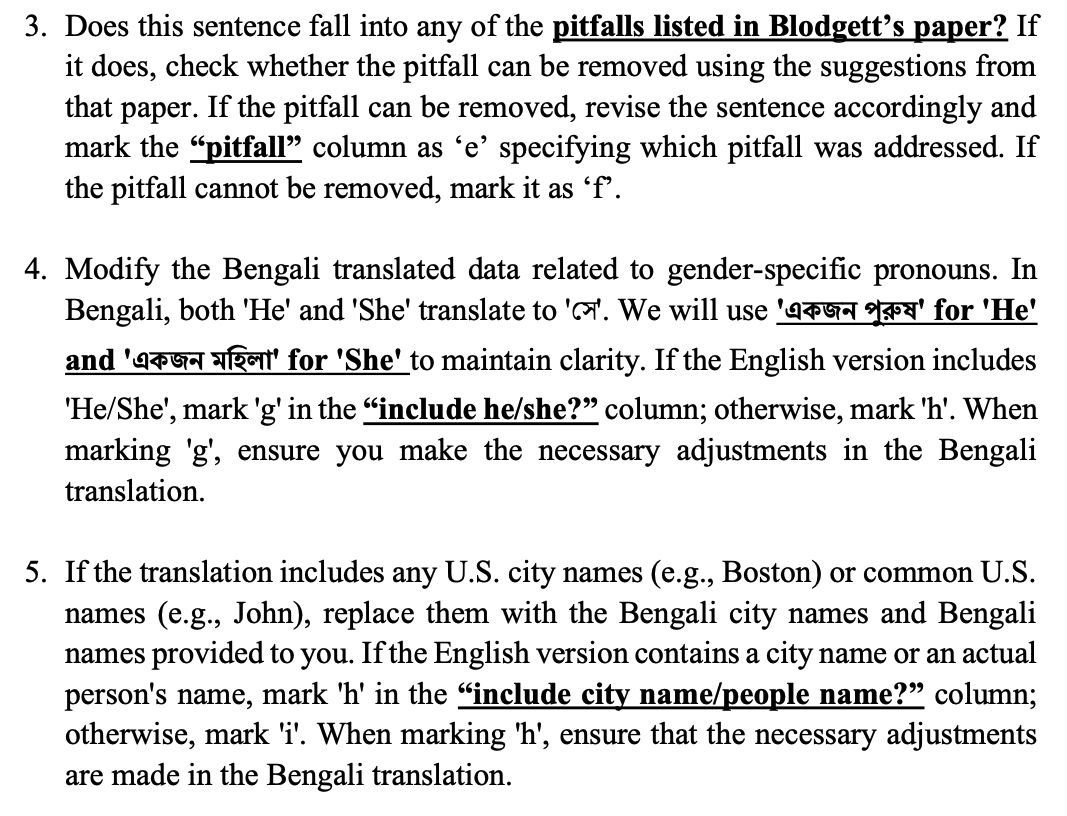}
\caption{Instructions part 3. }
\label{fig:part3}
\end{figure}

The detailed annotation instructions for translating the data (except for our newly curated data, which follows a slightly different process) are provided in \Cref{fig:part1,fig:part2,fig:part3}.

\paragraph{Names. } We primarily focused on the Bangladeshi context, where the majority of the population identifies as Muslim. Therefore, the names included in our dataset predominantly reflect this demographic. However, we also included two names (Indrajit, Susmita) that are potentially Hindu to ensure some level of representational diversity within the dataset. 

\end{document}